\documentclass[journal]{IEEEtran}
\ifCLASSINFOpdf
\else
\fi
\usepackage[hidelinks]{hyperref}   % to hide link borders
\usepackage{hyperref}
\usepackage{subcaption}
\usepackage{epstopdf}
\usepackage{graphicx}
\usepackage{amsfonts}
\usepackage{enumerate}
\usepackage{varioref}
\usepackage{longtable}
\usepackage{graphicx}
\usepackage{amssymb}
\usepackage{chngpage}
\usepackage{multirow}
\usepackage{multirow}
\usepackage{amsthm}
\usepackage{amsmath}
\usepackage{caption}
\usepackage{nccmath}  % to left align an equation
\usepackage[table]{xcolor}  % for color in table
\usepackage{algorithm}% http://ctan.org/pkg/algorithms               ----> % for algorithm (pseudo-code)
\usepackage{algpseudocode}% http://ctan.org/pkg/algorithmicx 		 ----> % for algorithm (pseudo-code)

\usepackage{threeparttable}

\begin{document}

\title{Recognizing Involuntary Actions from 3D Skeleton Data Using Body States}

\author{Mozhgan Mokari,
		Hoda Mohammadzade,		
		Benyamin Ghojogh

\thanks{Mozhgan Mokari's e-mail: \href{mailto:mokari_mozhgan@ee.sharif.edu}{mokari\_mozhgan@ee.sharif.edu}}
\thanks{Hoda Mohammadzade's e-mail: hoda@sharif.edu}
\thanks{Benyamin Ghojogh's e-mail: \href{mailto:ghojogh_benyamin@ee.sharif.edu}{ghojogh\_benyamin@ee.sharif.edu}}
\thanks{All authors are with Department of Electrical Engineering, Sharif University of Technology, Tehran, Iran}}

%\markboth{To be Submitted to IEEE Transactions on Systems, Man, and Cybernetics - Part C: Applications and Reviews}%

\maketitle

\begin{abstract}
Human action recognition has been one of the most active fields of research in computer vision for last years. Two dimensional action recognition methods are facing serious challenges such as occlusion and missing the third dimension of data. Development of depth sensors has made it feasible to track positions of human body joints over time. This paper proposes a novel method of action recognition which uses temporal 3D skeletal Kinect data. This method introduces the definition of body states and then every action is modeled as a sequence of these states. The learning stage uses Fisher Linear Discriminant Analysis (LDA) to construct discriminant feature space for discriminating the body states. Moreover, this paper suggests the use of the Mahalonobis distance as an appropriate distance metric for the classification of the states of involuntary actions. Hidden Markov Model (HMM) is then used to model the temporal transition between the body states in each action. According to the results, this method significantly outperforms other popular methods, with recognition rate of 88.64\% for eight different actions and up to 96.18\% for classifying fall actions.
\end{abstract}


\IEEEpeerreviewmaketitle
%%%%%%%%%%%%%%%%%%%%%%%%%%%%%%%%%%%%%%%%%
\section{Introduction}

\IEEEPARstart{S}{ince} last two decades, human action recognition has drawn lots of attention from researches in computer vision and machine learning fields. In early attempts for action recognition, RGB video was used as input of recognition system. Various valuable methods and algorithms were proposed for recognizing actions and activities using RGB data. However, several problems exist in action recognition using RGB frames such as occlusion and different orientations of camera. Existence of other objects in addition to human bodies and the lack of information of the third dimension can be mentioned as other challenges in this category of methods \cite{16,17,18,19,20,21}.
In order to address these problems, methods for recognizing action from multiple views have been also introduced; however, they are typically very expensive in calculations and are not suitable for real time recognition \cite{22}. 

Due to the mentioned problems and by introducing 3D Kinect sensors in market, researchers started to work on 3D data for the purpose of action recognition. The Kinect sensor provides both depth and skeleton data in addition to capturing RGB frames. Different methods were proposed for either depth or skeleton data. 

\begin{figure}[!t]
\centering
\includegraphics[width=3.45in]{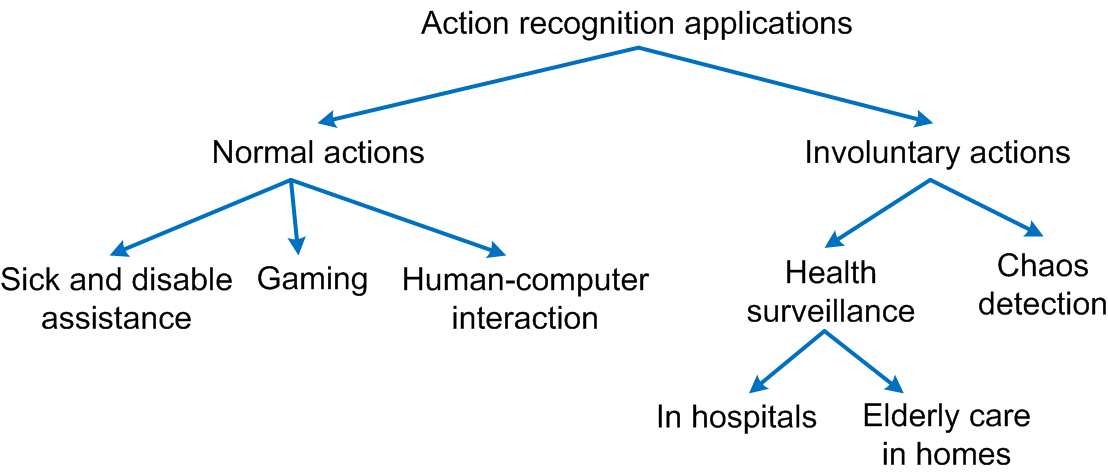}
\caption{Applications of human action recognition.}
\label{applications_fig}
\end{figure}

\begin{figure}[!t]
\centering
\includegraphics[width=2.5in]{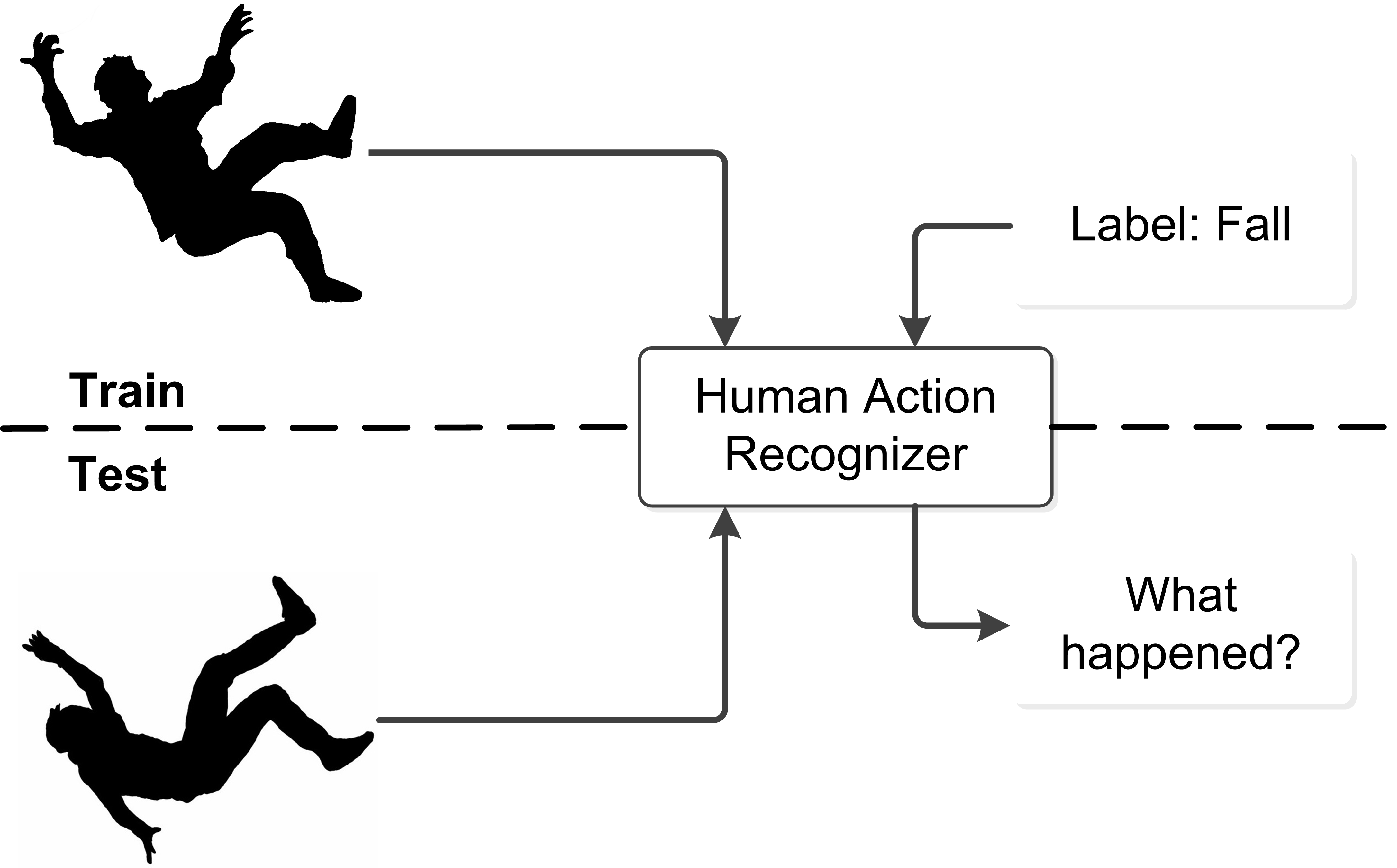}
\caption{A human action recognition system for fall detection.}
\label{HAR}
\end{figure}

Action recognition can have a variety kinds of applications. According to a vantage point, all actions can be categorized in one of two categories, i.e., normal actions and involuntary actions (see Fig. \ref{applications_fig}). Daily actions, actions for gaming, and interactions between human and computer can be considered as normal actions. On the other hand, involuntary actions can be seen in different situations, such as health surveillance and chaos detection. 
One of the most frequent involuntary actions is falling which can happen by patients in hospitals. Old people are also subject to dangerous falls, which if detected by surveillance systems for elderly cares can reduce serious injuries and fatalities. Another example of involuntary actions can be found in public surveillance systems which can detect chaos and abnormal actions in the crowd or interactions, and alert accordingly. 
Although, the proposed method in this work can be applied for both normal and involuntary actions, its focus is on involuntary actions. Figure \ref{HAR} depicts a human action recognition system used for fall detection.

This paper proposes a new method for human action recognition. In this method, an action is modeled as sequence of several body states. The definition of body states is introduced and then a method based on Fisher Linear Analysis (LDA) is proposed for their classification. The recognized body states are then fed to Hidden Markov Models (HMM) in order to model and classify actions. Mahalanobis distance is also utilized in this work in order to consider the different variations in performing involuntary actions. 
The results show that we can get high recognition accuracy even by using a a simple and linear classification method such as Fisher LDA for state recognition.

This paper is organized as follows. Section \ref{Related Work_section} reviews related work. Section \ref{Methodology_section} proposes the main algorithm of proposed method which includes modeling human body and action recognition using Fisher Linear Discriminant Analysis (LDA) and Hidden Markov Model (HMM). Section \ref{Experimental Results_section} introduces the utilized dataset and experimental results. Finally, Section \ref{Conclusion} concludes the article.

\section{Related Work}\label{Related Work_section}

According to the importance of action recognition and its large amount of applications, lots of different methods have been proposed in this field. Facing some challenges such as coverage of some part of body by others and introducing 3D methods, encouraged Li et al. \cite{1} to recognize the human's action by sampling the 3D points of depth image and creating an action graph. In this method, they model the position of human's body by projecting the contour of body shape onto the different planes and sampling them. Then, the state of human body is modeled with these bags of 3D points. The states are considered as nodes of a graph, modeling the action. Although this method is not robust against the changing of viewing angle and human's body scale, it has recognized 90 percent of actions and the error is halved compared to 2D methods.

Zhao et al. \cite{2} classified human's actions by utilizing information of RGB and depth image. They have obtained spatio-temporal interest points from RGB image and used combined descriptor of RGB and depth images.

By developing Kinect sensors and related software for tracking humans in images and detecting positions of human's joints in 3D space, several methods were proposed to recognize action using this information. One of these methods introduced Cov3DJ descriptor \cite{3} that separates different action classes by finding covariance matrix of positions of the joints during the action and used Support Vector Machine (SVM) for classification.

Reddy et al. \cite{4} recognized action by considering mean, minimum and maximum of position of joints as features and compared them to features obtained by using principle component analysis (PCA) on position of joints. In a similar research, Martnez-Zarzuela et al. \cite{5} tried to recognize actions by taking a sequence of positions of joints as a signal and extracting the five first Fast Fourier Transform (FFT) components as a feature vector, which is fed into a neural network. However, this method did not perform very well for complex actions that involve different parts of body.

As different actions involve different joints, Anjum et al. \cite{6} selected important and effective joints in the training level, according to the type of action. In their proposed method, each action is determined by three joints. Results showed that this method performs better with less information, but joints should be selected in training for each action; therefore, extending this algorithm for other new actions is time consuming and difficult.

Haynh et al. \cite{7} proposed a new method more robust to human scale and changes of position. It was performed by categorizing joints into three classes of stable, active and highly active joints and using angles of 10 important joints and vectors connecting moving joints to stable joints. Their method performed better than using only raw position of joints.

Xia et al. \cite{8} used middle and side hip joints to extract a histogram of position of other joints, which is used as feature. They reduced the dimension of feature vector using Linear Discriminant Analysis (LDA) and used K-means method to cluster the feature vectors. Each cluster constitutes a visual word. Each action is determined as a time sequence of these visual words and modeled by Hidden Markov Model (HMM). Results showed that this method has partially overcome challenges such as different lengths of actions and the same action done in different ways and view angles.

Papadopoulos et al. \cite{9} obtained orientation of body using the positions of shoulders and hip joints and there by, extracted orthogonal basis vectors for each frame. Therefore, a new space is constructed for every person according to its orientation of body. According to these vectors and the new space, the spherical angles of joints are used instead of position of joints.
Using angles instead of position of joints, made method more robust against human's body scale and changes in the shape of body. This method also uses energy function to overcome the challenge of same actions done by opposite hands or feet.

Although there are lots of proposed methods for action recognition, but many problems and challenges are still unsolved. This paper tries to face some of them such as different distributions of actions in statistical feature space, in order to improve the existing methods, especially for involuntary actions.

\section{Methodology}\label{Methodology_section}

In order to recognize actions, at the first step, the actions should be modeled in an appropriate way. Modeling actions depends on various facts such as application, types of actions and method of classification.
One of the most important applications of action recognition is online recognition which performs in real time. This article goals this type of recognition. In this category, the action should be modeled so that the model can be updated during completion of action and finally recognize the type of performed action. Therefore, in this article, each action is supposed to be a sequence composed of several states of body.

\begin{figure}[!t]
\centering
\includegraphics[width=2in]{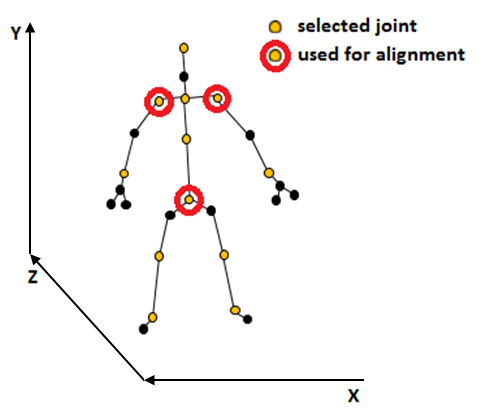}
\caption{Selected joints out of available joints in the skeletal data. The joints used for alignment are also shown.}
\label{joints}
\end{figure}

In the next step, position of joints in the 3D space are utilized in order to model the state of body. The position of joints are prepared by the output of Kinect sensor. The skeleton consists of several joints, which are 25 joints for the dataset used for experiments in this paper. Several joints are, however, so close to each other without any important difference in movements; therefore, their information are almost redundant. With respect to the actions addressed in this paper, merely 12 important joints, which are right and left ankles, right and left knees, right and left wrists, right and left shoulders, head, middle spine, hip and spine shoulder are selected out of the skeleton. Position of spine base (hip) and right and left shoulders are used for alignment in order to correctly describe the state of body in different persons and runs. The selected joints and also joints required for alignment are shown in Fig. \ref{joints}. State modeling and skeleton alignment are detailed in the following.

\subsection{Modeling State of Body}

In order to model and describe the state of body, a proper descriptor should be created. This descriptor models the action as a time sequence of states and tries to recognize the action. 
Different locations and orientations of body in the frames forces the need to aligning the skeleton.

\subsubsection{Aligning Skeleton}

\begin{figure}[!t]
\centering
\includegraphics[width=3.45in]{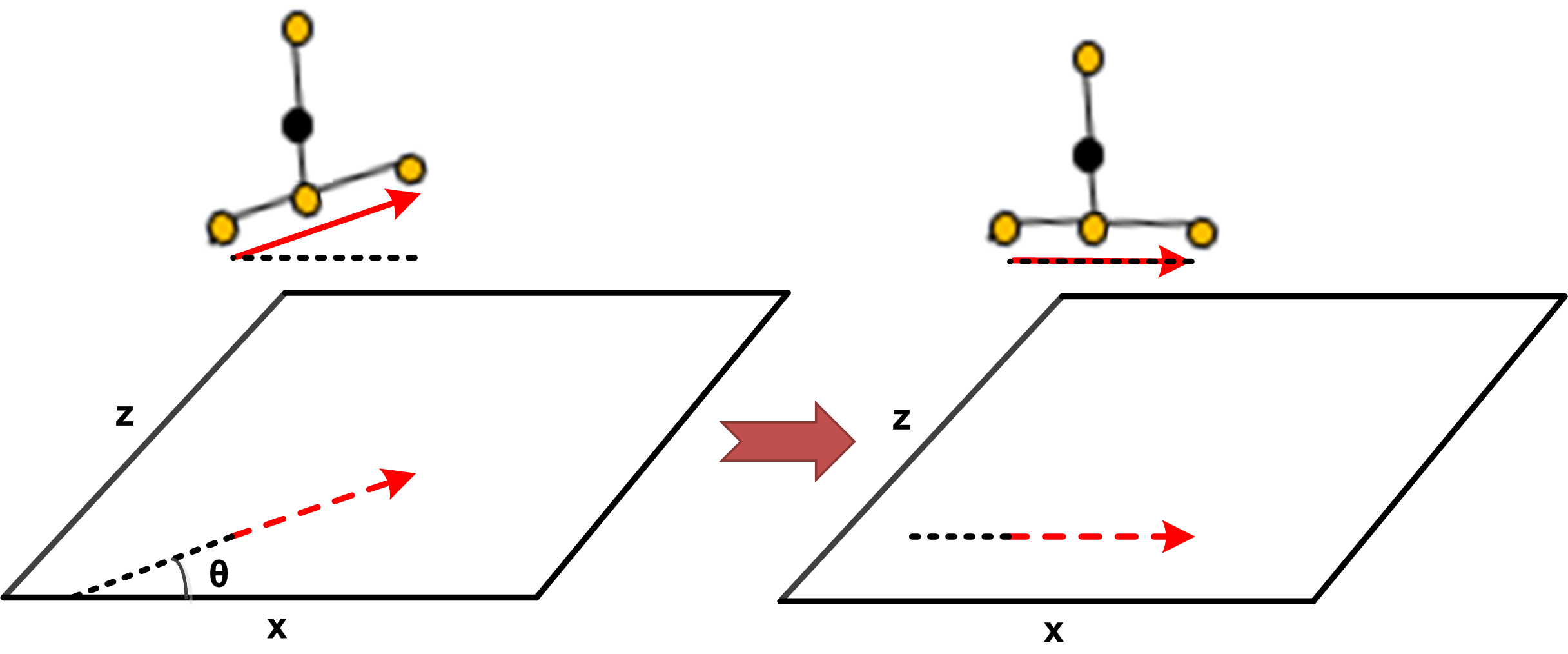}
\caption{Alignment of skeleton using the left and right shoulders to cancel the orientation of skeleton.}
\label{Alignment}
\end{figure}

As already mentioned, 12 joints positions are used in 3D space, in order to describe the state of body. 
In order to cancel the location of body skeleton, the position of hip joint is subtracted from the position of all joints. This is performed for every frame in the sequence.

Moreover, different orientations of skeleton or camera makes recognizing similar states difficult and even wrong. Thus, in order to cancel different orientations of body skeletons, the body is rotated around $y$ axis making the projection of the vector connecting the left and right shoulder onto the $xz$ plane parallel to the $x$ axis. By performing this rotation, the skeleton directly faces the camera.
This procedure is illustrated in Fig. \ref{Alignment}. The axes can be seen in Fig. \ref{joints}.
In literature, alignment of skeleton is also common, but the methods or used joints for that might differ. For example, in \cite{8}, left and right hip joints are utilized rather than shoulder joints for alignment.

\subsubsection{Creating feature vector}

To determine the state of body in each frame, proper feature vectors are required. Three joints out of the 12 joints are used for alignment and the remaining nine joints are used to create the feature vectors. Fisher Linear Discriminant Analysis (LDA) \cite{14} is utilized for having proper features. In Fisher LDA method, the dimension of feature vector is reduced to $C-1$, where $C$ is the number of states. In LDA, the within- ($S_w$) and between-class ($S_b$) scatter matrices are

\begin{equation}
S_w = \sum_{i=1}^C \sum_{x_k \in X_i} (x_k - \mu_i) (x_k - \mu_i)^T,
\end{equation}

\begin{equation}
S_b = \sum_{i=1}^C N_i (\mu_i - \mu) (\mu_i - \mu)^T,
\end{equation}
in order to minimize the within class covariance and maximize the between class covariance \cite{10,11}, where $\mu_i$ and $\mu$ denote the mean of $i^{th}$ state and mean of means, respectively. 
The Fisher projection space is created by the eigenvectors of $S_w^{-1}S_b$.
By its projection in this space, the feature vector $F$ for an input skeleton state is obtained.

After projection onto Fisher space, the obtained feature vectors are located relative to each other such that similar and various states respectively fall close and apart. By this fact, recognition of states becomes available.

\begin{figure}[!t]
\centering
\includegraphics[width=2in]{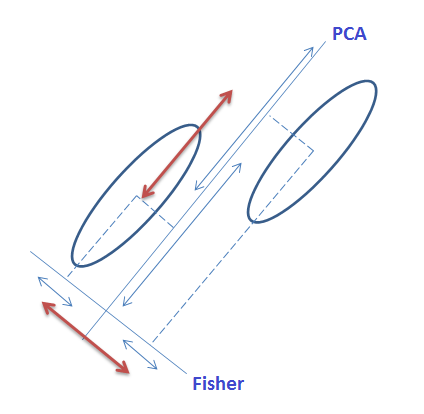}
\caption{An example of Fisher and PCA directions.}
\label{Fisher_and_PCA}
\end{figure}

There are also other methods for feature reduction which can be used for classification. One of the most popular methods of this category is Principle Component Analysis (PCA) \cite{10,11}. However, PCA method cannot always classify the data as well as LDA does. As an example, suppose that the distribution of classes are similar to that depicted in Fig. \ref{Fisher_and_PCA}. In this example, the Fisher LDA direction is perpendicular to the direction of PCA. As is obvious in this figure, Fisher LDA tries to lower within-class variance and maximize between-class variance in order to classify them. 

The final feature vector is used for training and testing the state of body. The action will be defined as a time sequence of multiple specific states. The state of body is recognized in test phase, by finding the minimum distance as described in the following sections.

\subsubsection{Finding the minimum distance}

In every frame denoted as $f$, the state of body should be recognized. For reaching this goal, the distances between feature vector $F$ of this frame and the means of all feature vectors of states are found. The minimum distance determines the state of the frame $f$. If $\widetilde{F}_i$ denotes the mean of feature vectors of the $i^{th}$ class, the state is found as,

\begin{equation}
\text{state}(f) = \text{arg }\underset{i}{\text{min }} d(F,\widetilde{F}_i),
\end{equation}
where $d$ is the distance measurement function which can be one of the two followings:

\begin{itemize}
\item \textit{Euclidean Distance:} One of the popular methods for calculating the distance of two vectors is Euclidean distance, which is used as one of the distance methods in this article. The function of Euclidean distance can be formulated as,

\begin{equation}
d(F,\widetilde{F}_i) = \sqrt{\sum_{j}^{} (F_j - \widetilde{F}_{ij})^2},
\end{equation}
where $F_j$ and $\widetilde{F}_{ij}$ are the $j^{th}$ component of $F$ and $\widetilde{F}_i$, respectively. 
 
\item \textit{Mahalanobis Distance:} As the minimum distance from the means of states is used for recognizing the state, defining the distance has much important influence on the accuracy of recognition. Therefore, the distribution of final feature vectors in the feature space should be considered and the distance measurement should be defined according to it.

As the dimension of final feature (Fisher) vectors is $C-1$ and the states of utilized dataset are categorized into eight classes, the dimension of final feature vectors is seven.
The dimensions higher than two or three cannot be illustrated; although, the distribution of feature vectors can be analyzed in higher dimensions by calculating their covariance matrices.
The first two directions of Fisher space are used for illustration of distribution of each state.
Figure. \ref{Distributions} illustrates the training samples projected onto the space constructed by the first two Fisher directions. 
As shown in this figure, distribution of projected states are various in different directions.

\begin{figure}[!t]
\centering
\includegraphics[width=3.45in]{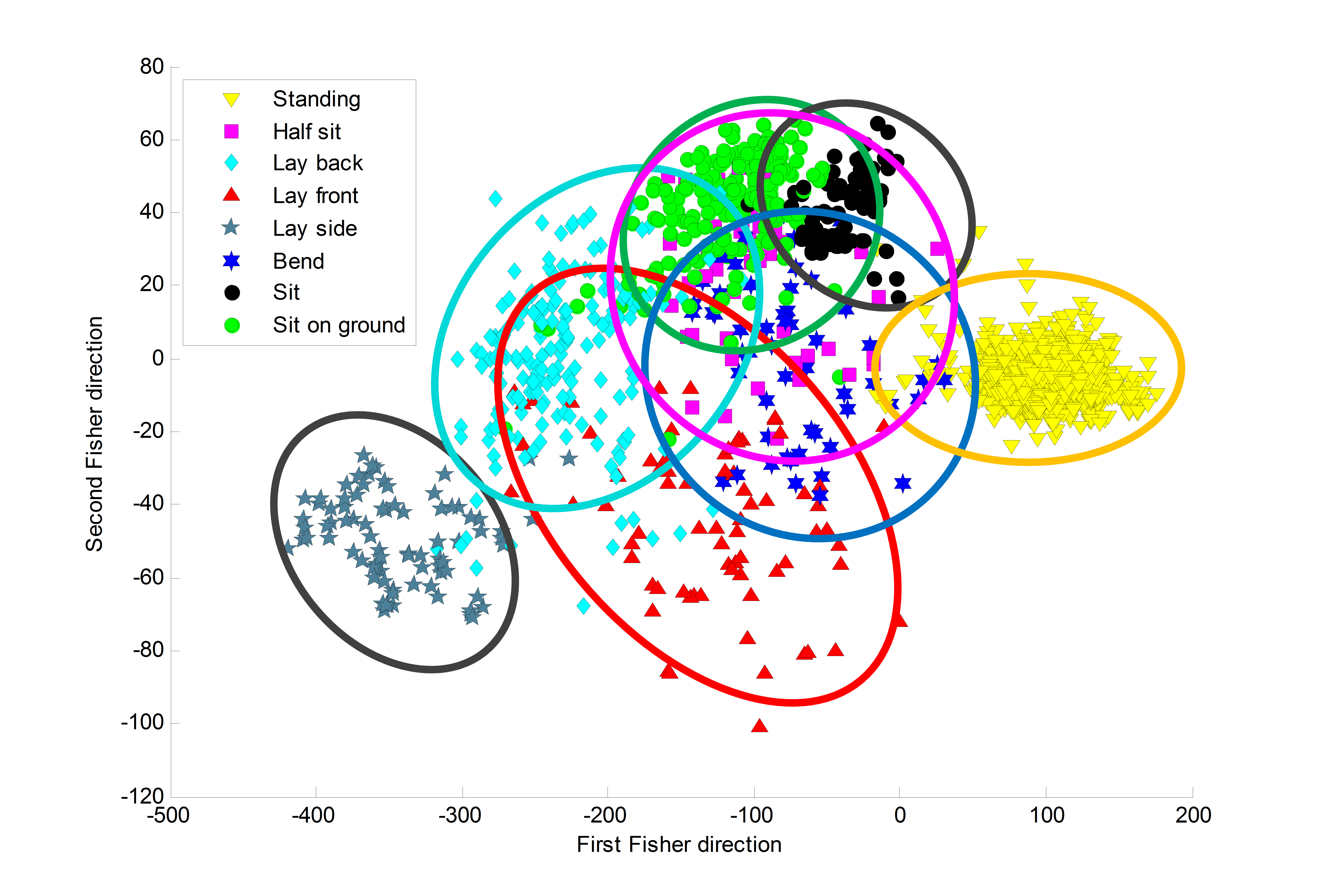}
\caption{Projection of samples of states onto Fisher space. As can be seen, the states have different distributions.}
\label{Distributions}
\end{figure}

The more differently people perform an action containing a state, the more distributed the state would be.
The more distributed states usually are between the states with low variance, during the completion of an involuntary action. For instance, as shown in Fig. \ref{Distributions}, after projection on constructed Fisher space, normal states such as standing and sit are less distributed than the states occurred in abnormal actions, such as lay front and lay back.
In order to handle the challenge of different distributions of projected states, a distance measurement function should be used which considers the distributions rather than Euclidean distance. 

Mahalanobis distance considers variances of distributions in its calculation.
The Mahalanobis distance is calculated as,

\begin{equation}\label{mahalanobis_formula}
d(F,\widetilde{F}_i) = \sqrt{ (F - \widetilde{F}_i)^T S^{-1} (F - \widetilde{F}_i) },
\end{equation}
where $S$ denotes the covariance matrix of the feature vectors of a class.
As is obvious in equation \eqref{mahalanobis_formula}, the covariance matrix $S$ is modeling a weight for each class according to its distribution. Therefore, the importance of distance in a particular dimension is considered in calculating the distances. In other words, the distance in a direction with smaller variance is less valuable, yielding to $S^{-1}$ in the equation.

Mahalanobis distance is actually an extension to the standard deviation from the mean, in multi-dimensional space. Experiments reported in following sections, show outperformance of this distance in comparison with Euclidean distance.
 
\end{itemize}

\subsection{Classyfing Actions Using Hidden Markov Model}\label{HMM_section}

As previously mentioned, every action can be modeled as a sequence of consequent states. After recognizing states of body using Fisher LDA, Hidden Markov Model (HMM) is utilized in this work to classify actions.

\begin{figure}[!t]
\centering
\includegraphics[width=2.5in]{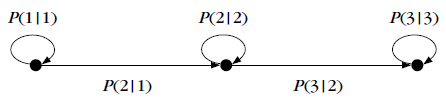}
\caption{A three state HMM model \cite{12}.}
\label{HMM}
\end{figure}

For every action, a separate HMM is used. 
HMM uses several sequences of states as observations and models the particular pattern that occurs during an action. Note that the state of body should not be confused with state of HMM, here. 
In HMM, each state have a probability of happening and also there exist transitional probabilities between states.
For instance, a three state HMM and its transitional probabilities are illustrated in Fig. \ref{HMM} \cite{12}.

In order to decrease computational cost of the algorithm, the frame per second rate has been reduced by down sampling.
After constructing a HMM for each action, an unknown sequence is recognized by feeding it to each HMM. 
After feeding the test sequence of frames to all trained HMMs, every HMM outputs a probability of occurrence for that sequence. The maximum probability determines the action of that sequence.

For each action, the sequences that are used for training HMM are set to have the same lengths (number of states). However, this equalization of the length of sequences is not related to the length and speed of action performed by different persons. This equalization is performed by manually repeating the last state done by the person; so that the total number of states of all actions become similar.

The advantage of HMM, in this work, is that it considers solely the dynamic of sequence and is not sensitive to various paces and lengths of actions. In fact, the number of repeats of a state does not significantly impact the recognition of that action. This gives the method robustness to different paces of actions and even different speeds of one action.  
For instance, there exist sequences of lengths 75 frames upto 463 frames with different speeds of actions in TST fall dataset \cite{13,15}, and these sequences have been successfully handled and recognized by this method.

\begin{figure}[!t]
\centering
\includegraphics[width=2.5in]{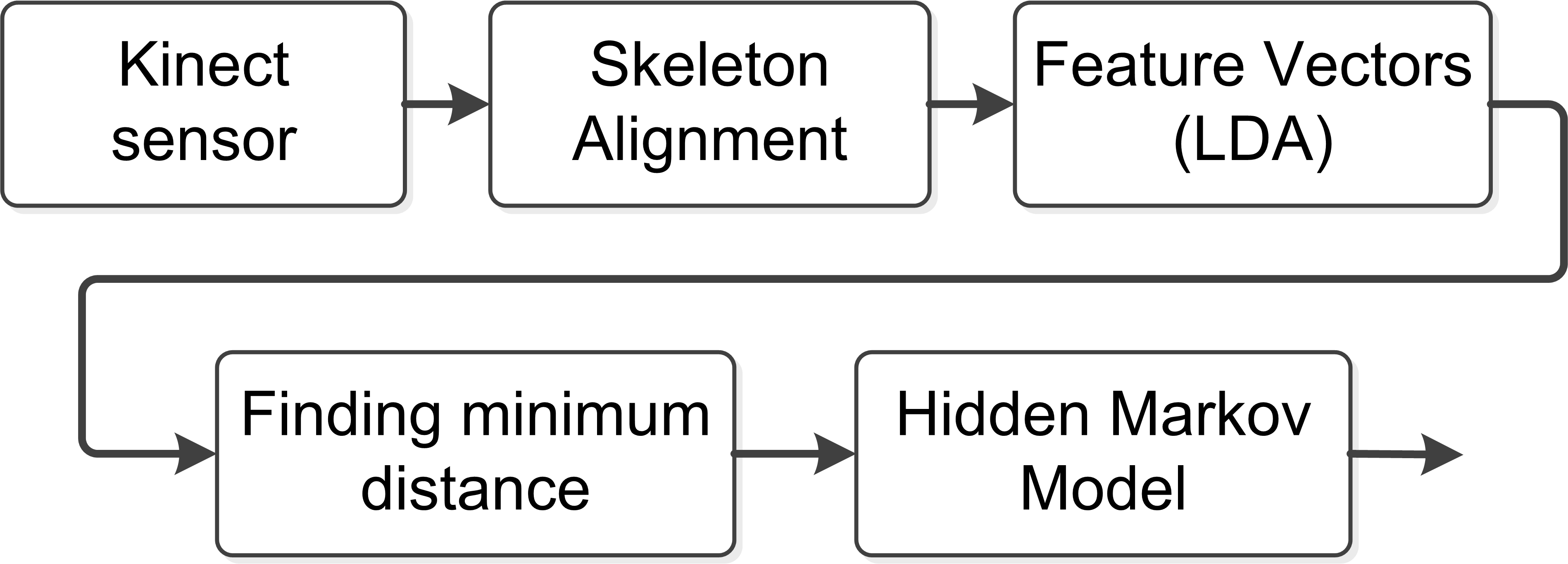}
\caption{The overall structure of proposed framework.}
\label{The_Structure}
\end{figure}

The overall structure of the proposed framework is summarized in Fig. \ref{The_Structure}. 

\section{Experimental Results}\label{Experimental Results_section}

To examine the proposed method, TST Fall Detection dataset \cite{13} is used. The details of this dataset are explained in next section following by modeling the actions. At the end, the results of experiments are presented.

\begin{figure}[!t]
\centering
\begin{subfigure}[b]{0.2\textwidth}
\centering
\includegraphics[width=1in]{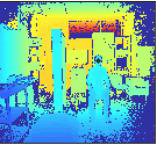} 
\caption{Sit}
\label{1}
\end{subfigure}
\begin{subfigure}[b]{0.2\textwidth}
\centering
\includegraphics[width=1in]{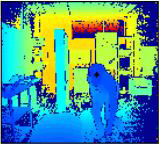} 
\caption{Grasp}
\label{1}
\end{subfigure}
\begin{subfigure}[b]{0.2\textwidth}
\centering
\includegraphics[width=1in]{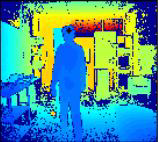} 
\caption{Walk}
\label{1}
\end{subfigure}
\begin{subfigure}[b]{0.2\textwidth}
\centering
\includegraphics[width=1in]{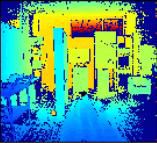} 
\caption{Lay down}
\label{1}
\end{subfigure}
\begin{subfigure}[b]{0.2\textwidth}
\centering
\includegraphics[width=1in]{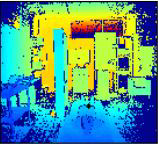} 
\caption{Fall front}
\label{1}
\end{subfigure}
\begin{subfigure}[b]{0.2\textwidth}
\centering
\includegraphics[width=1in]{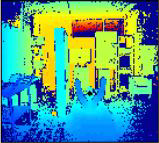} 
\caption{Fall back}
\label{1}
\end{subfigure}
\begin{subfigure}[b]{0.2\textwidth}
\centering
\includegraphics[width=1in]{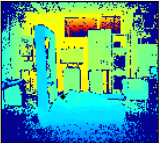} 
\caption{Fall side}
\label{1}
\end{subfigure}
\begin{subfigure}[b]{0.2\textwidth}
\centering
\includegraphics[width=1in]{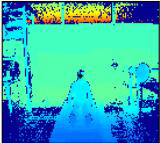} 
\caption{End up sit}
\label{1}
\end{subfigure}
\caption{An example of actions in TST dataset \cite{13}.}
\label{actions}
\end{figure}

\subsection{Dataset}

TST Fall Detection dataset \cite{13, 15} is used for verifying the effectiveness of this method. There are two main categories of actions in this dataset, i.e., daily living activities and fall actions. 
11 different persons perform every action for three times.
The daily living activities are sit, lying down, grasp and walk and the fall actions are falling front, back, side and end up sit.

This dataset has prepared information of 3D position of joints and depth data obtained by the Kinect sensor V2, which is more accurate than previous Kinect sensors. Only the skeletal data of this dataset is used in this work for experiments. 
This work concentrated mostly on the involuntary actions such as falling down which exist sufficiently in this dataset.

As previously depicted in Fig. \ref{HAR}, one of the possible goals in human action recognition is surveillance application especially for controlling elderly or patient people.
The main goal of detecting involuntary actions and improvements of Kinect V2 encouraged this work to use the mentioned dataset.

In this dataset, the significant challenge is existence of abnormal and involuntary actions such as falling, in addition to the normal actions. As fall actions are performed involuntarily, different states and conditions from normal actions appear in different people. The challenge is to develop a method which can detect and analyze these actions and also recognize them from each other. Hence, the methods proposed for recognizing normal actions, may not necessarily perform as well for fall actions.

In addition, existing methods for recognizing fall actions cannot be utilized; because they have concentrated on features such as speed and acceleration which distinguish fall actions from all normal actions. These features do not discriminate the normal actions from each other and therefore do not help recognizing the actions in general.

Several samples of depth images of actions in TST dataset are shown in Fig. \ref{actions}.

\subsection{Recognition of states}

\begin{figure*}[!t]
\centering
\begin{subfigure}[b]{0.3\textwidth}
\centering
\includegraphics[width=2in]{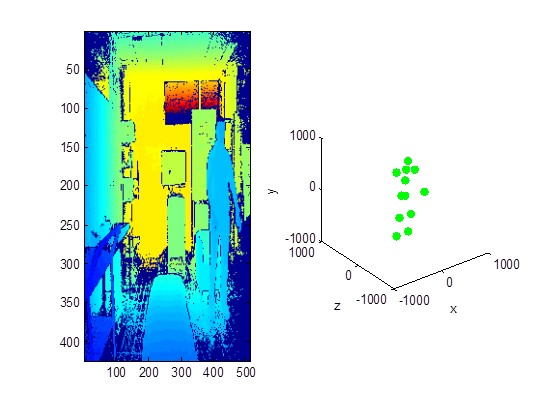} 
\caption{Stand}
\label{1}
\end{subfigure}
\begin{subfigure}[b]{0.3\textwidth}
\centering
\includegraphics[width=2in]{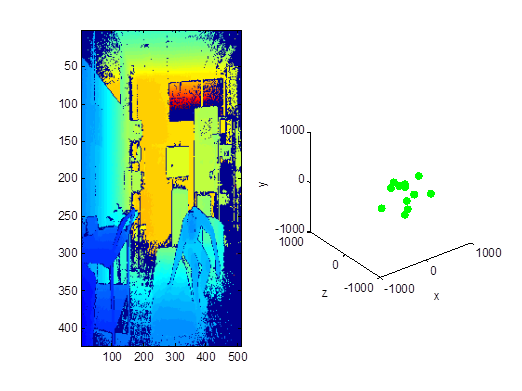} 
\caption{Crouching}
\label{1}
\end{subfigure}
\begin{subfigure}[b]{0.3\textwidth}
\centering
\includegraphics[width=2in]{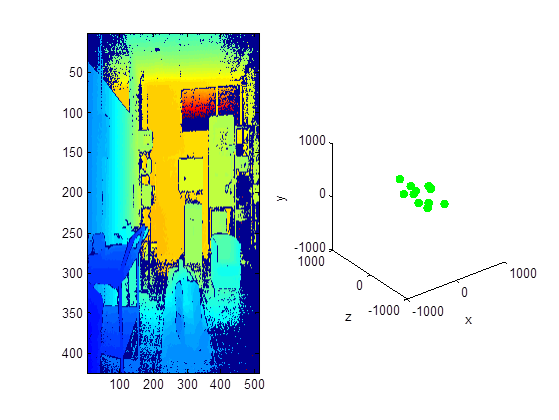} 
\caption{Lay down back}
\label{1}
\end{subfigure}
\begin{subfigure}[b]{0.3\textwidth}
\centering
\includegraphics[width=2in]{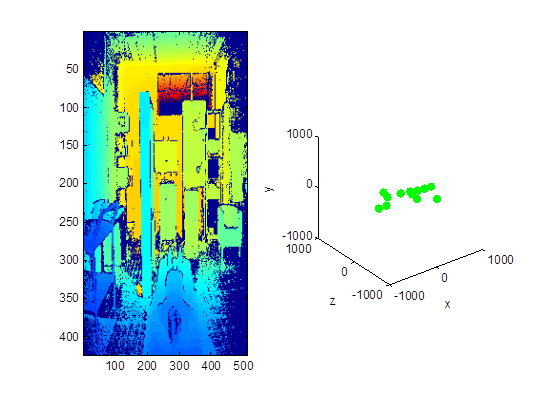} 
\caption{Lay down front}
\label{1}
\end{subfigure}
\begin{subfigure}[b]{0.3\textwidth}
\centering
\includegraphics[width=2in]{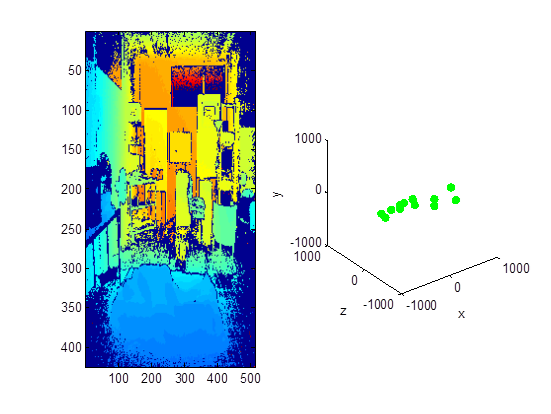} 
\caption{Lay down side}
\label{1}
\end{subfigure}
\begin{subfigure}[b]{0.3\textwidth}
\centering
\includegraphics[width=2in]{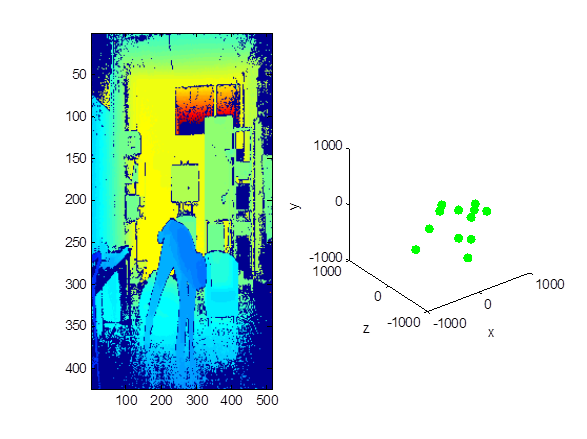} 
\caption{Bend}
\label{1}
\end{subfigure}
\begin{subfigure}[b]{0.3\textwidth}
\centering
\includegraphics[width=2in]{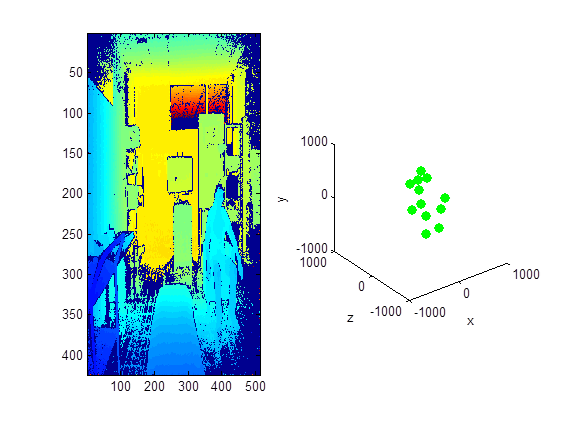} 
\caption{Sit on chair}
\label{1}
\end{subfigure}
\begin{subfigure}[b]{0.3\textwidth}
\centering
\includegraphics[width=2in]{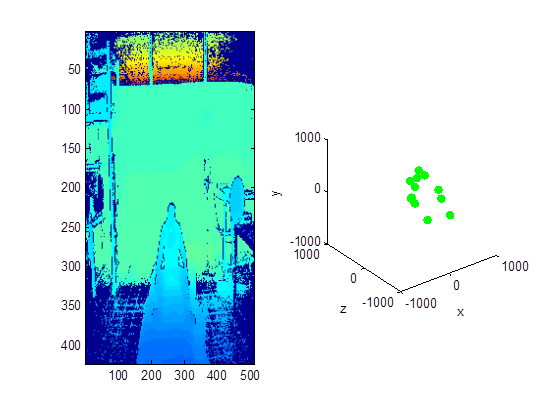} 
\caption{Sit on ground}
\label{1}
\end{subfigure}
\caption{An example of the selected states.}
\label{states}
\end{figure*}

In the dataset only the actions are labeled and labeling states should be performed manually. According to the actions, eight different states are chosen and labeled to be used in LDA. 
The chosen states should include the main states of actions in the dataset and should not contain unnecessary states which are close to other states.
The chosen states are stand, crouching, lay back, lay front, lay side, bend, sit on chair and sit on ground. An example of each state is shown in Fig. \ref{states}.

\begin{figure}[!t]
%\begin{minipage}{\textwidth}
\centering
\begin{subfigure}[b]{0.5\textwidth}
\centering
\includegraphics[width=3.45in]{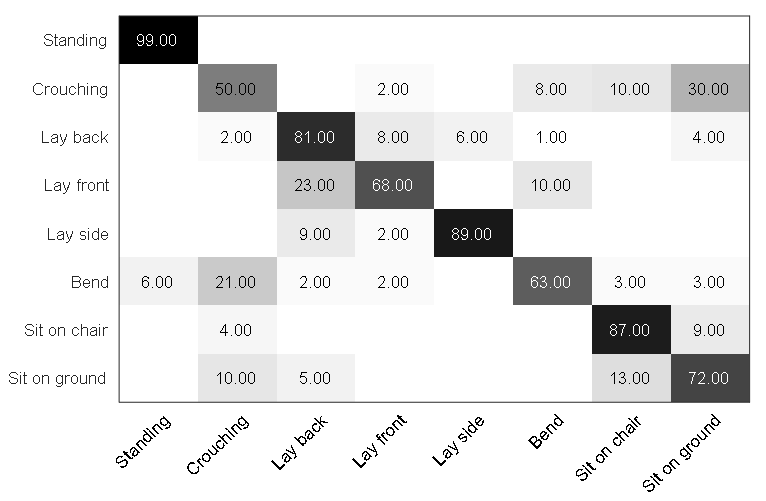} 
\caption{Euclidean}
\end{subfigure}
\begin{subfigure}[b]{0.5\textwidth}
\centering
\includegraphics[width=3.45in]{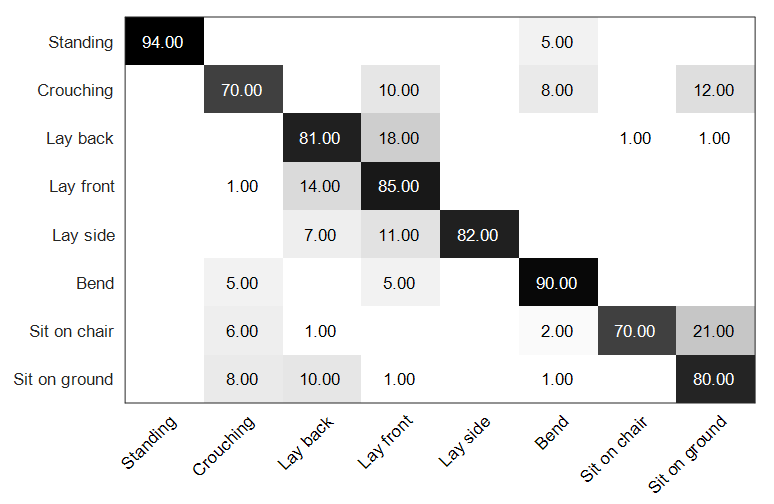} 
\caption{Mahalanobis}
\end{subfigure}
\caption{Confusion matrix of states.}
\label{Confusion_matrix_states}
%\end{minipage}
\end{figure}

The \textquotedblleft leave one subject out\textquotedblright \text{} cross validation is used for the experiments. In each iteration, the entire samples of a person is considered as test samples and the samples of other subjects are used for training system. This type of cross validation is fairly difficult because the system does not see any sample from the test subject in training phase. The state recognition experiment is repeated using both of the distance methods and the results are listed in Table \ref{state_result}.

\begin{table}[!t]
%\begin{minipage}{\textwidth}
\renewcommand{\arraystretch}{1.3}  %%% each row size
\caption{Correctness rate of recognizing state of body}
\label{state_result}
\centering
\begin{tabular}{l | c c}
\hline
\hline
\textbf{State} & \textbf{Euclidean} & \textbf{Mahalanobis}\\
\hline
Stand & 99.38\% & 94.26\% \\
\hline
Crouching & 50.00\% & 70.00\% \\
\hline
Lay back & 80.71\% & 81.22\% \\
\hline
Lay front & 67.50\% & 85.00\% \\
\hline
Lay side & 88.89\% & 82.22\% \\
\hline
Bend & 62.90\% & 90.32\% \\
\hline
Sit & 86.87\% & 69.70\% \\
\hline
Sit on ground & 72.15\% & 79.91\% \\
\hline
\hline
Total & \textbf{76.03}\% & \textbf{81.57}\% \\
\hline
\hline
\end{tabular}
%\end{minipage}
\end{table}

Table \ref{state_result} shows that the Mahalanobis distance outperforms the Euclidean distance in general. 
As was expected, the recognition rates of crouching, lay front and bend have been improved significantly using Mahalanobis distance. The reason is that the variances of training data for these states are huge and this fact is not taken into account when Euclidean distance is used.

It is worth noting that using the Mahalanobis distance, the recognition rate of bend state has been improved at the cost of reducing the recognition rate of stand state. Closer look at Fig. \ref{Distributions} reveals that there exists an overlapping region of distributions between the two states. Euclidean distance which does not consider the distribution of classes, mostly recognizes the overlapping region as the stand state. On the other hand, the Mahalanobis distance mostly recognizes this region as the bend state, because the variance of standing state is much less than bend. This fact can also be seen from the confusion matrix of different states which are depicted in Fig. \ref{Confusion_matrix_states} for both distances.

\subsection{Action Recognition \& Comparison}

In the last step, an action is divided into several states in order to be described as a sequence of states. Each state in this sequence is recognized by projecting into the LDA space and utilizing a distance measure. 
Then, the probability of each HMM (note that there is an HMM for each specific action) generating the input sequence of states is calculated and maximum probability determines the recognized action. The number of states parameter for HMM's (note that the state here is different from the state of body) affects the recognition performance.
Therefore, different number of states were tested for HMM's in this work and were compared to each other. Results of three different numbers of state for HMM's are reported in Table \ref{HMM_states}. The experiments of this table are performed with Mahalanobis distance. 
As was expected according to the nature of states and actions in the TST Fall dataset \cite{13,15}, HMM's with three states perform better and hence, the number of states for HMM's is considered to be three in this work. It is worth nothing that, combination of optimum number of states for each action was also considered, but the experiments showed that use of a constant number of states for all HMM's results in better performance.  

In this article, the proposed method is compared with the method of Xia et al. \cite{8} which has received  considerable attention in literature \cite{23,24,25,26} and has been used for comparison in very recent methods \cite{27,28,29,30,31}. Note that all the above methods has experimented with datasets created using an older version of Kinect sensor and not containing involuntary actions.

For implementing method \cite{8} and fairly comparing it with the proposed method using the TST dataset, several necessary adjustments were performed in its settings. 
First, for LDA, the states are labeled in the same way as in the proposed method. Second the number of states for HMM's was chosen to be three, according to the actions of the dataset. Third the best number of clusters for histogram was experimented to be eight, which conforms with the number of classes of states in the proposed method.

\begin{table}[!t]
%\begin{minipage}{\textwidth}
\renewcommand{\arraystretch}{1.3}  %%% each row size
\caption{Effect of number of states of HMM on the recognition rate}
\label{HMM_states}
\centering
\begin{tabular}{l | c c c}
\hline
\hline
\textbf{Action} & \textbf{2 states} & \textbf{3 states} & \textbf{4 states}\\
\hline
Sit & 87.88\% & 90.91\% & 90.91\% \\
\hline
Grasp & 90.91\% & 90.91\% & 87.88\% \\
\hline
Walk & 93.94\% & 93.94\% & 93.94\% \\
\hline
Lay & 84.85\% & 96.97\% & 90.91\% \\
\hline
Front & 84.85\% & 81.82\% & 81.82\% \\
\hline
Back & 84.85\% & 84.85\% & 78.79\% \\
\hline
Side & 81.82\% & 81.82\% & 81.82\% \\
\hline
End up sit & 84.85\% & 87.88\% & 84.85\% \\
\hline
\hline
Total & 86.74\% & \textbf{88.64\%} & 86.36\% \\
\hline
\hline
\end{tabular}
%\end{minipage}
\end{table}

\begin{table}[!t]
%\begin{minipage}{\textwidth}
\renewcommand{\arraystretch}{1.3}  %%% each row size
\caption{Comparison of results of our method and method \cite{8} for TST dataset}
\label{action_result}
\centering
\begin{threeparttable}[b]
\begin{tabular}{l | c c c}
\hline
\hline
\textbf{Action} & \textbf{Euclidean} & \textbf{Mahalanobis} & \textbf{\cite{8}}\\
\hline
Sit & 84.85\% & 90.91\% & 81.82\% \\
\hline
Grasp & 96.97\% & 90.91\% & 84.85\% \\
\hline
Walk & 100\% & 93.94\% & 90.91\% \\
\hline
Lay & 75.76\% & 96.97\% & 90.91\% \\
\hline
Front & 54.54\% & 81.82\% & 48.49\% \\
\hline
Back & 69.70\% & 84.85\% & 66.67\% \\
\hline
Side & 81.82\% & 81.82\% & 69.70\% \\
\hline
End up sit & 69.70\% & 87.88\% & 33.33\% \\
\hline
\hline
Total & 79.16\% & \textbf{88.64\%} & 70.83\% \\
\hline
\hline
\end{tabular}
%\end{minipage}
\end{threeparttable}
\end{table}

\begin{figure}[!t]
%\begin{minipage}{\textwidth}
\centering
\begin{subfigure}[b]{0.5\textwidth}
\centering
\includegraphics[width=3.45in]{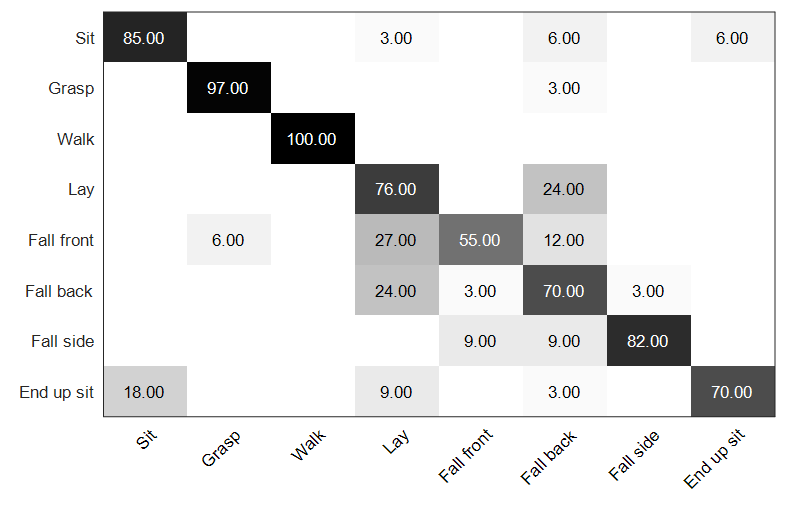} 
\caption{Euclidean}
\end{subfigure}
\begin{subfigure}[b]{0.5\textwidth}
\centering
\includegraphics[width=3.45in]{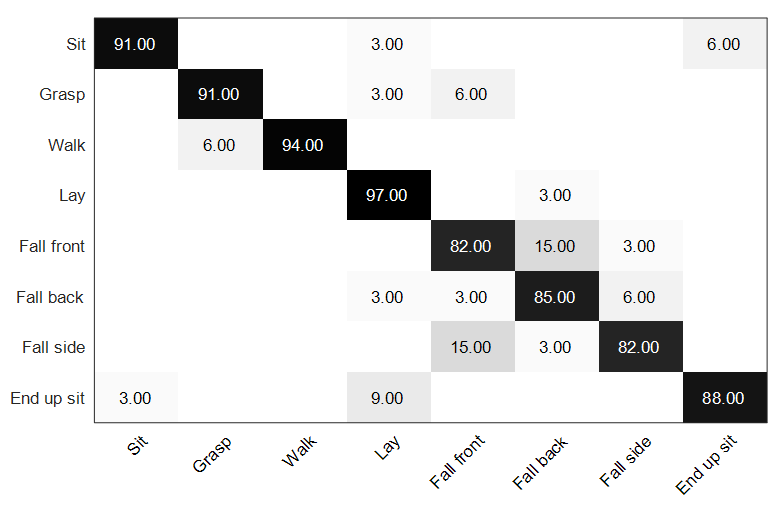} 
\caption{Mahalanobis}
\end{subfigure}
\caption{Confusion matrix of actions.}
\label{Confusion_matrix}
%\end{minipage}
\end{figure}

Results are listed in Table \ref{action_result}. The proposed method using both distance methods are compared with the method of Xia et al. \cite{8}. Results reveal that in all actions, the proposed method using each of the two distance measures outperforms the method \cite{8}. Although method \cite{8} has utilized LDA and clustering methods in preparing data for training HMM; it has made several states so close to each other by using a histogram concept and has increased error. As an example, in fall actions the angular positions of joints are much similar and the use of histogram ignores their differences.

Using Mahalanobis distance has significantly enhanced the performance of recognition, especially in fall actions. In other words, improving the performance of recognizing difficult involuntary states such as crouching and lay front, has improved the total recognition rate. As mentioned before, the main reason of this fact is that the intrinsic variance of states are considered in Mahalanobis distance.

The confusion matrix of action recognition is reported in Fig. \ref{Confusion_matrix}. This matrix shows that the actions that are similar to each other, are sometimes confused and wrongly recognized. Actions such as falling on front, side and back are sometimes confused with each other; because their distribution (and therefore their behavior) are similar and wider than others, as is obvious in Fig. \ref{Distributions}. These actions are referred to as abnormal actions. In some cases such as anomaly detection in actions, this wrong recognition might not matter. Based on application, considering all abnormal and all normal actions as two different high-level groups, the recognition rate will improve from 88.64\% to 96.18\%. 
And as can be seen in Table \ref{fall_result}, false alarm rate has also been significantly reduced. It indicates that the possibility of wrongly recognizing a normal action as fall action is considerably low.

\begin{table}[!t]
%\begin{minipage}{\textwidth}
\renewcommand{\arraystretch}{1.3}  %%% each row size
\caption{Comparison of results, considering all abnormal actions to be fall event}
\label{fall_result}
\centering
\begin{threeparttable}[b]
\begin{tabular}{l | c c c}
\hline
\hline
& \textbf{Euclidean} & \textbf{Mahalanobis} & \textbf{\cite{8}} \\
\hline
\textbf{\shortstack{Recognition Rate \\ (true positive rate)}} & 78.78\% & \textbf{96.18}\% & 77.27\% \\
\hline
\textbf{\shortstack{Specificity Rate \\ (true negative rate)}} & 90.15\% & \textbf{96.21}\% & 90.90\% \\
\hline
\textbf{\shortstack{False Alarm Rate \\ (false positive rate)}} & 9.15\% & \textbf{3.78}\% & 9.09\% \\
\hline
\hline
\end{tabular}
%\end{minipage}
\end{threeparttable}
\end{table}

\section{Conclusion}\label{Conclusion}

A new action recognition method was proposed in this paper, especially for recognizing the actions that have some complexities such as different forms of falling. Since this method uses feature vectors with low dimension and does not have big computational overhead, can be used in real time purposes. Experiments showed that this method outperforms the other methods especially for unusual actions.

In this method, a feature vector is created for determining the state of body in each frame, using the Kinect data and Fisher LDA method. 
Actions are classified and recognized by feeding the sequence of recognized states of body to HMMs. Because of using HMM, this method is robust to different paces and lengths of actions.
Moreover, the Mahalanobis distance is utilized for considering the variance of data in order to enhance the recognition rate.

Data was preprocessed by skeleton alignment, to make the algorithm robust against the orientation of camera. As future work, the angles between the joints can be used instead of their positions in order to get more robustness. In addition, recognizing more complex and longer actions can be considered as a future work.

\section{Acknowledgment}

This work was supported by a grant from Iran National
Science Foundation (INSF).

% or


\appendices
%%%%%\section{Proof of the First Zonklar Equation}
%%%%%Appendix one text goes here.

%%%%%\section{}
%%%%%Appendix two text goes here.

%%%%%\section*{Acknowledgment}

%%%%%%%%%%%%%%%The authors would like to thank...

\ifCLASSOPTIONcaptionsoff
  \newpage
\fi



%\bibliographystyle{IEEEtran}


\hfill \break

\begin{IEEEbiographynophoto}{Mozhgan Mokari}
received her BSc degree in electrical engineering from Amirkabir University of Technology (Tehran Polytechnic), Tehran, Iran, in 2014. She also received her MSc degree in electrical engineering from Sharif University of Technology, Tehran, Iran, in 2016. She is currently studying for PhD of electrical engineering in Sharif University of technology. Her research interests are machine learning, computer vision and signal processing.
\end{IEEEbiographynophoto}

\begin{IEEEbiographynophoto}{Hoda Mohammadzade}
received her BSc degree from Amirkabir University of Technology (Tehran Polytechnic), Iran, in 2004, the MSc degree from the University of Calgary, Canada, in 2007, and the PhD degree from the University of Toronto, Canada, in 2012, all in electrical engineering. She is currently an assistant professor of electrical engineering at Sharif University of Technology, Tehran, Iran.  Her research interests include signal and image processing, computer vision, pattern recognition, biometric systems, and bioinformatics.
\end{IEEEbiographynophoto}

\begin{IEEEbiographynophoto}{Benyamin Ghojogh}
obtained his first and second BSc degrees in electrical engineering (Electronics and Telecommunications fields) from Amirkabir University of technology, Tehran, Iran, in 2015 and 2017 respectively. He also received his MSc degree in electrical engineering (Digital Electronic Systems field) from Sharif University of technology, Tehran, Iran, in 2017. One of his honors is taking the second rank of Electrical Engineering Olympiad of Iran in 2015. His research interests include machine learning and computer vision.
\end{IEEEbiographynophoto}


%\newpage



%\vfill


\end{document}